\documentclass[runningheads]{llncs}
\usepackage[T1]{fontenc}
\usepackage{comment}
\usepackage{amsmath}
\usepackage{graphicx}
\begin{document}
\title{An Adaptive Neuro-Controller Developed for a Prosthetic Hand Wrist }
%
%
\author{Shifa Sulaiman*\orcidID{0000-0002-5330-0053} \and
Francesco Schetter \and
Mohammad Gohari\and
Fanny Ficuciello} 
\authorrunning{S. Sulaiman et al.}
%
\institute{Department of Information Technology and Electrical Engineering, Università degli Studi di Napoli Federico II, Claudio, 21, 80125 Napoli, Italy \\
\email{*ssajmech@gmail.com}}
\maketitle              
\begin{abstract}
The significance of employing a controller in prosthetic hands cannot be overstated, as it plays a crucial role in enhancing functionality and usability of prosthetic hands. 
This paper introduces an adaptive neuro-controller specifically developed for a tendon-driven soft continuum wrist of a prosthetic hand. Kinematic and dynamic modellings of the wrist are carried out using the Timoshenko beam theory. A Neural Network (NN) based strategy is adopted to predict required motor currents to manipulate tendons of the wrist from errors in deflections of the wrist section. Timoshenko beam theory is used for computing required tendon tension from the input motor current. Comparison of the adaptive neuro-controller with other similar controllers is carried out to analyse the performance of the proposed controller. We also included simulation studies and experimental validations to illustrate the proposed controller's efficacy. 

\end{abstract}
\section{INTRODUCTION}
 
Soft robotic prostheses \cite{ref1,4} represent a cutting-edge innovation that offers individuals with limb impairments a more comfortable and natural range of motion compared to conventional rigid prostheses. Constructed from flexible materials, these soft prostheses emulate the dynamics of human muscles and tendons, allowing for greater adaptability and responsiveness to the user's specific requirements. 
The incorporation of soft continuum sections enables intricate movements, making them suitable for a wide array of applications. Elastic wires integrated with these soft continuum segments function as tendons, providing flexibility, lightweight characteristics, cost-effectiveness, and the ability to endure substantial tensile forces. 
The importance of employing an adaptive neuro-controller for manipulating soft continuum sections assists in enhancing system performance through real-time learning and adaptation. This advanced control strategy integrates neural network (NN) principles with adaptive control techniques, allowing it to adjust to dynamic environments and varying system parameters effectively. By utilizing the adaptive neuro controller, systems can achieve improved accuracy and stability, as it continuously refines its control actions based on feedback from the environment. This capability is particularly beneficial in complex and nonlinear systems where traditional control methods may struggle. Furthermore, the adaptive neuro controller can facilitate the optimization of performance metrics, leading to more efficient operation and reduced energy consumption, thereby making it a valuable tool in various applications, including robotics, automotive systems, and industrial automation. The major contributions of the presented work are as follows:
\begin{itemize}
    \item Modelling of a soft continuum wrist based on the Timoshenko beam theory.
    \item Development of a hybrid NN and Timoshenko model-based controller for the wrist motions.
    \item A comparison study of the proposed controller with similar controllers to demonstrate the advantages of the proposed controller.
    \item Experimental validations proving the effectiveness of the proposed controller during real-time implementations with reduced computational effort. 
\end{itemize}
The structure of this paper is organized as follows: Section 2 reviews existing literature concerning the advancement of controllers for soft continuum mechanisms. Modeling methodologies are detailed in Section 3. Adaptive neuro-controller strategy is illustrated in Section 4. Section 5 showcases the simulation and experimental findings. Finally, Section 6 concludes the study.

\section{State of art}
Neuro-controllers facilitate real-time decision-making and problem-solving capabilities, which are essential in various applications, including robotics, automation, and artificial intelligence \cite{2,3}. Their ability to process vast amounts of data and recognize patterns allows for improved control strategies, ultimately leading to more responsive and effective systems in diverse fields such as healthcare, manufacturing, and transportation. 
Braganza \textit{et al.} \cite{ref3} introduced a controller for continuum robots that incorporated an NN feed-forward element to address dynamic uncertainties. Experimental findings with the OCTARM, a soft extensible continuum manipulator, demonstrated that integrating the NN feed-forward component enhanced the controller's performance. An adaptive neuro-fuzzy control system designed for managing a flexible manipulator with a varying payload was demonstrated in \cite{ref4}. The proposed controller integrated a fuzzy logic controller (FLC) within a feedback loop, alongside two dynamic recurrent NNs in the forward path. Specifically, a dynamic recurrent identification network (RIN) was utilized to ascertain the output of the manipulator system, while a dynamic recurrent learning network (RLN) was employed with learning the weighting factor of the fuzzy logic. 
In \cite{ref5}, another FLC approach was introduced along with an adaptive neuro fuzzy inference strategy (ANFIS) to manage the input displacement of an innovative adaptive-compliant gripper. 

One of the major constraints of continuum manipulators concerning their structure is that even minor adjustments in the actuated lengths can lead to considerable variations in the position of the end effector. So controllers should be fast with less computational effort. Nowadays, reinforcement learning (RL) strategies are employed to develop motion control systems that enhance the responsiveness of soft robots. A continuous-time Actor-Critic framework designed for tracking tasks involving continuum 3D soft robots that are influenced by Lipschitz disturbances was presented in \cite{ref6}. This approach utilized a reward-based temporal difference method, which facilitated learning through an innovative discontinuous adaptive mechanism for the Critic neural weights. Ultimately, the integration of the reward and the Bellman error approximation enhanced the adaptive mechanism for the Actor neural weights. A model-free methodology for an open-loop position control of a soft spatial continuum arm, utilizing deep RL techniques was illustrated in \cite{ref7}. Deep-Q Learning with experience replay was used to facilitate the training of the system within a simulated environment.

 Melingui \textit{et al.} \cite{ref9} emphasized the necessity of implementing adaptive algorithms to address adverse effects. However, it has been also observed that merely employing adaptive control laws is inadequate due to the evolving nature of the robot model over time. An innovative adaptive control strategy known as the adaptive support vector regressor controller adopted for a position controller of a soft continuum robot was proposed in \cite{ref10}. The method utilized optimization learning techniques that provided global solutions while maintaining compact regressor sizes. These features enabled a faster convergence of the closed-loop system, consequently minimizing execution time. An adaptive NN control comprised of two subcontrollers developed for a bionic hand was presented in \cite{ref11}.  The first subcontroller managed the kinematics of the hand through a distal supervised learning approach. In contrast, the second subcontroller managed the kinetics of the hand utilizing adaptive neural control. Together, these subcontrollers enhanced the evaluation of the control architecture's stability while guaranteeing the convergence of Cartesian errors.
An NN based Adaptive controller strategy for a soft robotic arm was given in \cite{ref13}. The dynamics model of the arm was developed by integrating screw theory with Cosserat theory. Unmodeled dynamics within the system were taken into account, and an adaptive NN controller was formulated utilizing the back-stepping approach alongside a radial basis function NN.

A comprehensive literature review on the advancement of soft continuum robots has revealed that the existing generation of these robots faces multiple limitations, including inadequate kinematic and dynamic modeling techniques, suboptimal control schemes, and increased computational efforts. This paper outlines modelling methods and controller strategies employed to develop an adaptive neuro-controller for a soft wrist section attached to a prosthetic hand with faster response and reduced computational effort.  
\section{Mathematical model of a soft wrist section}
\begin{figure}[hbt!]
\centerline{\includegraphics[width=0.48\textwidth]{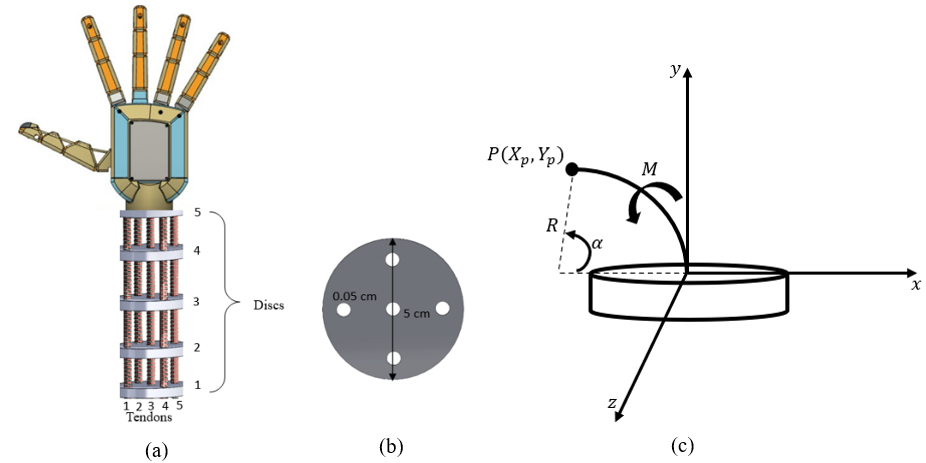}}
\caption{Soft wrist section (a) Conceptual design of wrist section attached to hand (b) Dimension of disc (c) Bending structure of wrist section}
\label{fig.1}
\end{figure}

\begin{figure}[hbt!]
\centerline{\includegraphics[width=0.25\textwidth]{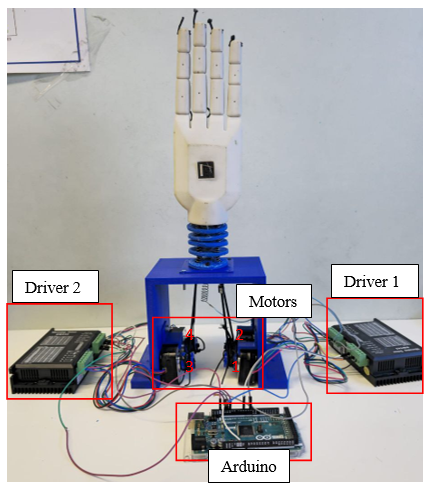}}
\caption{Fabricated model}
\label{fabricated}
\end{figure}

 The design of the proposed soft wrist segment, as detailed in \cite{ref15}, comprises five rigid discs, five springs, and five flexible tendons, as depicted in fig. \ref{fig.1}(a). The dimensions of the rigid discs utilized in this wrist segment are illustrated in fig. \ref{fig.1}(b). 
 The springs and tendons are integrated into the rigid discs and secured to a solid platform. By applying specific tensions to each tendon using a motor, the desired movements of the wrist segment can be achieved. The fabricated model of the wrist section integrated with a prosthetic hand is shown in fig. \ref{fabricated}.
 Variations in the tensions of the tendons lead to different bending moments acting on the soft wrist segment, thereby allowing its behavior to be modeled as a cantilever beam subjected to a bending moment. The positioning of the end effector in relation to the curvature of the wrist is determined through the principles of bending beam theory, as referenced in \cite{ref16}. Additionally, fig. \ref{fig.1}(c) provides a visual representation of the bending configuration of the soft wrist segment, with a length denoted as $L$, under the influence of an anti-clockwise moment, $M$.

\section{A neuro-controller developed for the soft wrist section }
\begin{figure*}[hbt!]
    \includegraphics[width=0.5\textwidth,height =0.6 in]{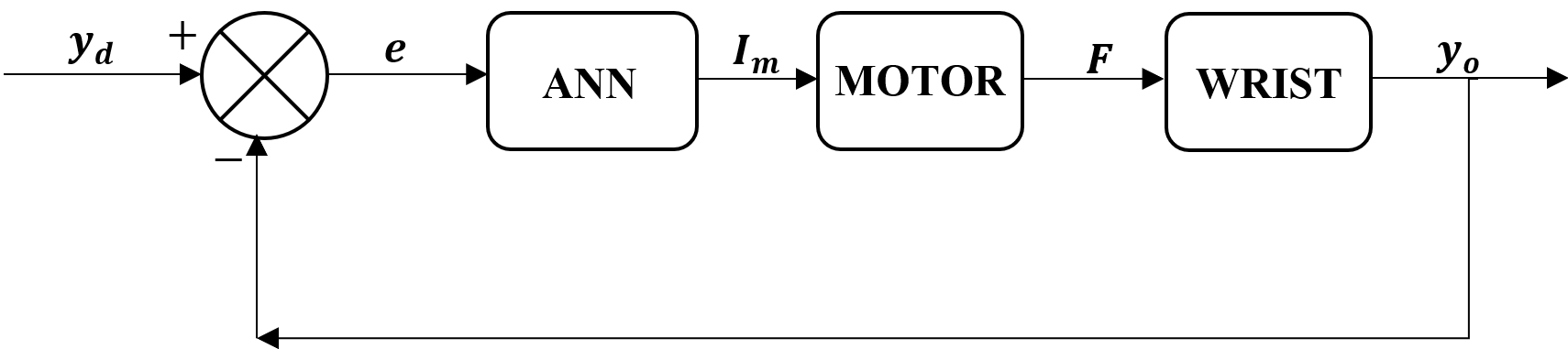}
    \includegraphics[width=0.45\textwidth,height =1.5 in]{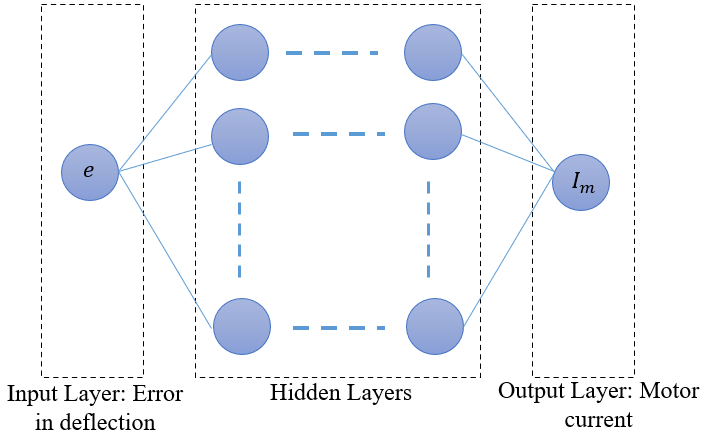}
    \caption{(a)Control scheme(b)ANN architecture}
   \label{ANN}
\end{figure*}

A neuro-controller (shown in fig. \ref{ANN}(a)) was developed for controlling the motions of the soft wrist section utilizing the Timoshenko beam theory. An artificial neural network (ANN) block was employed for predicting current values, $I_m$ as shown in fig. \ref{ANN}(b). 
The corresponding current values were fed to motors for providing required tendon tensions, $F$. The output deflections, $y_o$ obtained from the wrist section were used for computing errors, $e$ with respect to desired deflections, $y_d$ as given in equation \eqref{e} 
\begin{equation}
    e=y_d-y_o
    \label{e}
\end{equation}
These errors were received by the ANN block as the input and motor currents were given as the output. The motions of the motors attached to the tendons result in required tendon tensions to move the wrist section to desired deflections.
A conventional PID controller was initially employed for obtaining the input-output dataset by replacing ANN block shown in fig. \ref{ANN}(b). The error-current data pairs obtained from the PID controller scheme were used as datasets for training the ANN network in the neuro-controller scheme.  In the realm of control systems, the fundamental objective of employing a NN controller into the control scheme was to compute the control input, $u(t)$ that directed the system's state, $x(t)$ towards a desired target state, $x_{d}(t)$. The NN served to approximate the control law by leveraging the current state of the system, and in certain instances, it also incorporated the derivatives of the state variables. The neural network generated the control input, $u(t)$ based on the relationship given in equation \ref{NN}
\begin{equation}
    u(t) = \phi(w_2\phi(w_1x(t) + b_1) + b_2
    \label{NN}
\end{equation}
where $\phi$ and $x(t)$ are the activation functions of the NN and state vector of the system respectively. $w_1$ and $w_2$ are the weight matrices of the input-to-hidden and hidden-to-output layers respectively. $b_1$ and $b_2$ are the bias vectors of the hidden and output layers respectively. 
\section{Results and Discussions}
This study presents an adaptive neuro-controller designed for the soft wrist of a prosthetic hand. The kinematic and dynamic modellings of the wrist were conducted utilizing the Timoshenko beam theory. To study the performance of the proposed position controller during the motions of the soft wrist section carrying payload, simulation, comparison study, and experimentation were carried out.
\subsection{Simulation study}
\begin{figure*}
\includegraphics[width=0.45\textwidth, height = 2.7 in]{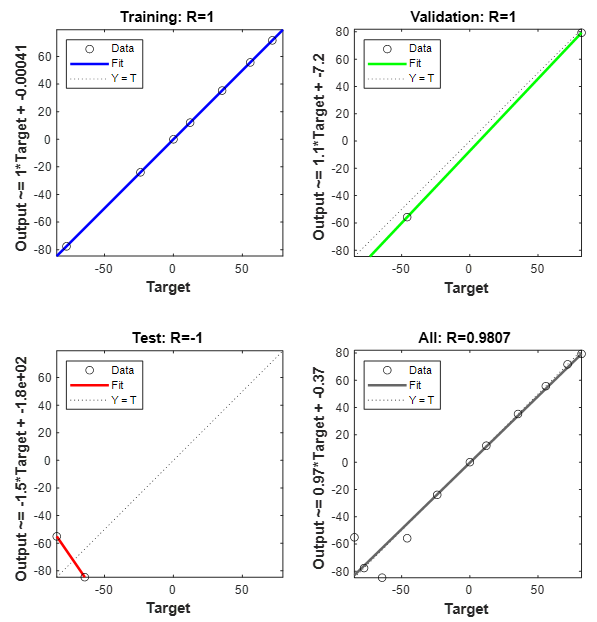}
\includegraphics[width=0.5\textwidth,height =2.71 in]{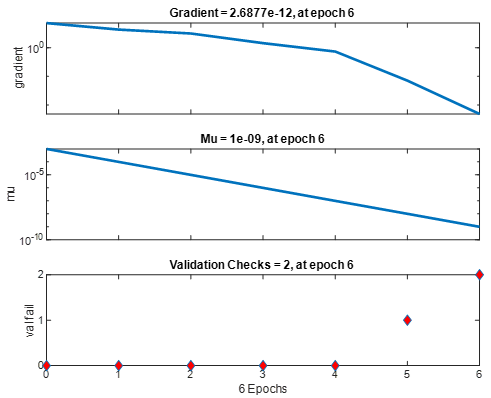}
    \caption{(a)Regression graphs(b)Gradient, momentum and validation check}
    \label{fig:reg}
\end{figure*}

\begin{figure}[hbt!]
    \centering
\includegraphics[width=0.45\textwidth]{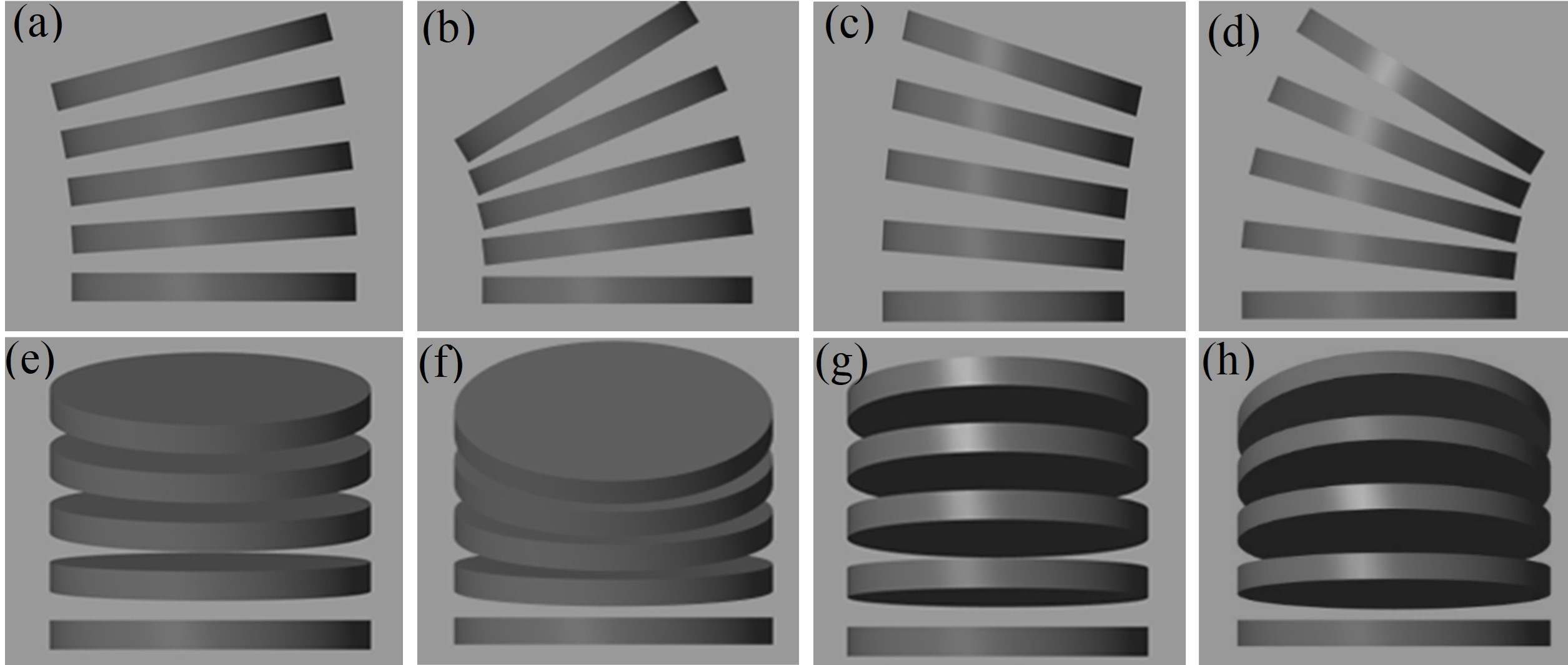}
   \caption{Motion of wrist (a) Radial-1 (b) Radial-2 (c) Ulnar-1 (d) Ulnar-2 (e) Flexion-1 (f) Flexion-2 (g) Extension-1 (h) Extension-2}
    \label{Motion of wrist}
\end{figure}

\begin{figure*}[hbt!]  
    \includegraphics[width=0.5\textwidth,height =1.9 in]{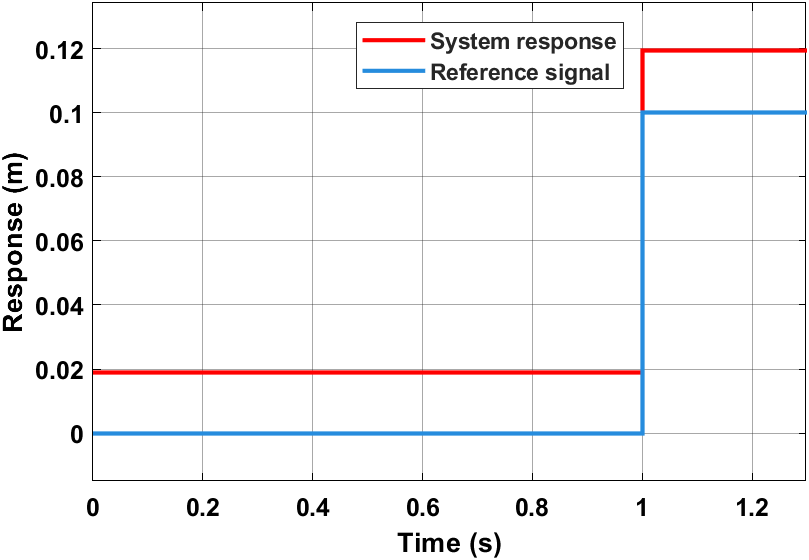}
     \includegraphics[width=0.5\textwidth,height =1.9 in]{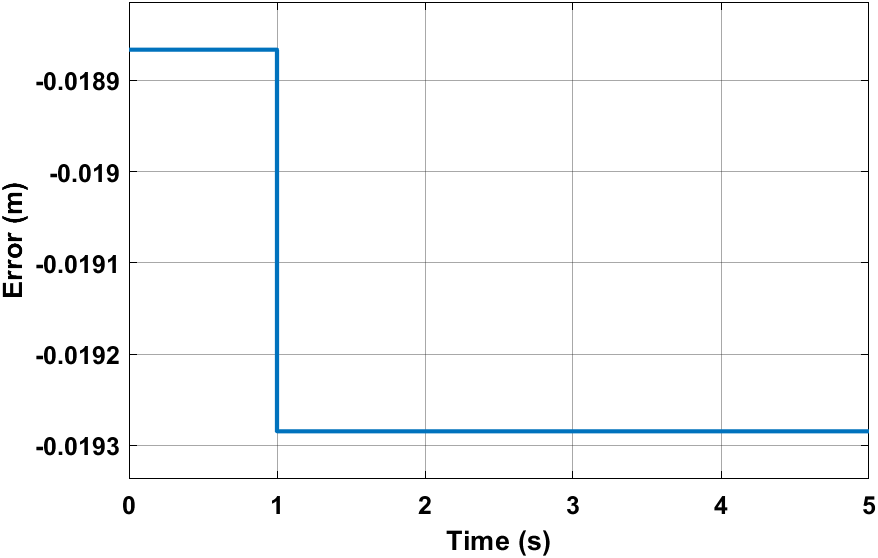}
    \caption{(a)Comparison of response with reference signal (b)Error in deflection during simulation}
   \label{response}
\end{figure*}

A PID controller was utilized to acquire the input-output dataset. The tuning of the PID controller for obtaining dataset was carried out using the crude tuning approach, resulting in the determination of the PID coefficients as $K_p=100$, $K_i=10$, and $K_d=100$. NN with feed-forward back propagation configuration was trained by using error signals as the input and obtained motor currents from the NN. NN was composed of 3 hidden layers with 5, 7, and 10 neurons in $1^{st}$, $2^{nd}$, and $3^{rd}$ layers respectively. Tansig function and Levenberg-Macquardt were used as the activation function and back propagation technique respectively. Regression scheme of the NN training is given in figs. \ref{fig:reg}. The regression scheme showcases an accuracy of 98.07 \% as evident from the fig. \ref{fig:reg}(a). 
Variations in gradient, momentum, and validation checks are shown in fig. \ref{fig:reg}(b).  Values of gradient and momentum (mu) were obtained as $2.67 \times 10^{-12}$ and $1 \times 10^{-9}$ respectively. Failed validation checks at epoch 6 were 2. The lower values of gradient, momentum, and validation failed checks at epoch 6 showcases the successful convergence of the NN. 
The parameters of the NN network are given in table \ref{table:1}. 
\begin{table}[hbt!]
\caption{NN based training parameters and results}
    \label{table:1}
    \centering
    \begin{tabular}{|c|c|}
    \hline
    Parameters &Value \\
    \hline
    Hidden layers &3 \\
        Gradient &$2.68 \times 10 ^{-12}$ \\
        Momentum & $1 \times 10 ^{-9}$ \\
        Learning rate & 0.0001       \\
        Accuracy  &0.98         \\
        Training loss &0.15            \\
        Validation loss &0.13    \\     
        \hline 
    \end{tabular}
\end{table}
The simulation of the control scheme was performed in Simulink, a MATLAB based software and carried out using a PC with an Intelcore i7 processor and 16 GB RAM. The wrist section can traverse trajectories in radial deviation, ulnar deviation, flexion, and extension directions as shown in figs. \ref{Motion of wrist} (a) - (h). 
The wrist segment was considered to be flexing from its original position, as illustrated in fig. \ref{Motion of wrist}, to a final bending angle of $30^{0}$ in ulnar deviation direction relative to disc 5 connected to the hand. Response of the system with respect to the reference step signal (reference deflection) is shown in fig. \ref{response}(a). The errors in deflections obtained during the simulation are shown in fig. \ref{response}(b).
Root Mean Square (RMSE), settling time, and steady state error were obtained as $0.05~m$, $1~s$, and $0.019~m$ respectively. 
\subsection{Comparison with other controllers}
The performance of the neuro-controller was compared with other controllers developed for the wrist section. Results obtained using a sliding mode controller (SMC) \cite{smc}, a model reference adaptive controller (MRAC) \cite{mrac}, and a geometric variable strain controller (GVSC) \cite{ref15} developed for the wrist section were compared with the results obtained using the proposed controller. RMSE, settling time and  steady state error were compared to analyse the performances of the controllers as given in table \ref{table:2}. 
\begin{table*}[hbt!]
\caption{Comparison study}
  \centering
    \begin{tabular}{|c|c|c|c|c|}
    \hline 
    Parameters & SMC \cite{smc} &MRAC\cite{mrac} &GVSC\cite{ref15} &Neuro-Controller\\
    \hline
     RMSE &$6.0 \times 10^{-3}$ &$1.2 \times 10^{-3}$  &$2.9 \times 10^{-2}$ &$5.0 \times 10^{-2}$  \\
    \hline 
   
 Settling time (s) &$1.5$ &$2.8$  &$3.18$ & $1$  \\
    \hline 
    Steady state error &$1.3 \times 10^{-3}~rad$ &$1.3 \times 10^{-3}~rad$ &$1.4 \times 10^{-2}~m$ &$1.9 \times 10^{-2}~m$   \\
    \hline 
     \end{tabular}\\
  \label{table:2}
\end{table*}
Neuro-controller converged faster and performed better in terms of settling time compared to other controllers. However, RMSE and steady state error values were higher compared to other controllers.
The performances showcased by SMC and MRAC were similar, since the steady state error was almost the same value. However, MRAC showcased lower RMSE and higher settling time compared to SMC. GVSC showcased higher settling time and lower RMSE and steady state errors values in comparison to the neuro-controller.

\subsection{Experimental validation}
\begin{figure}[hbt!]
    \centering
  \includegraphics[width=0.4\textwidth,height =1.2 in]{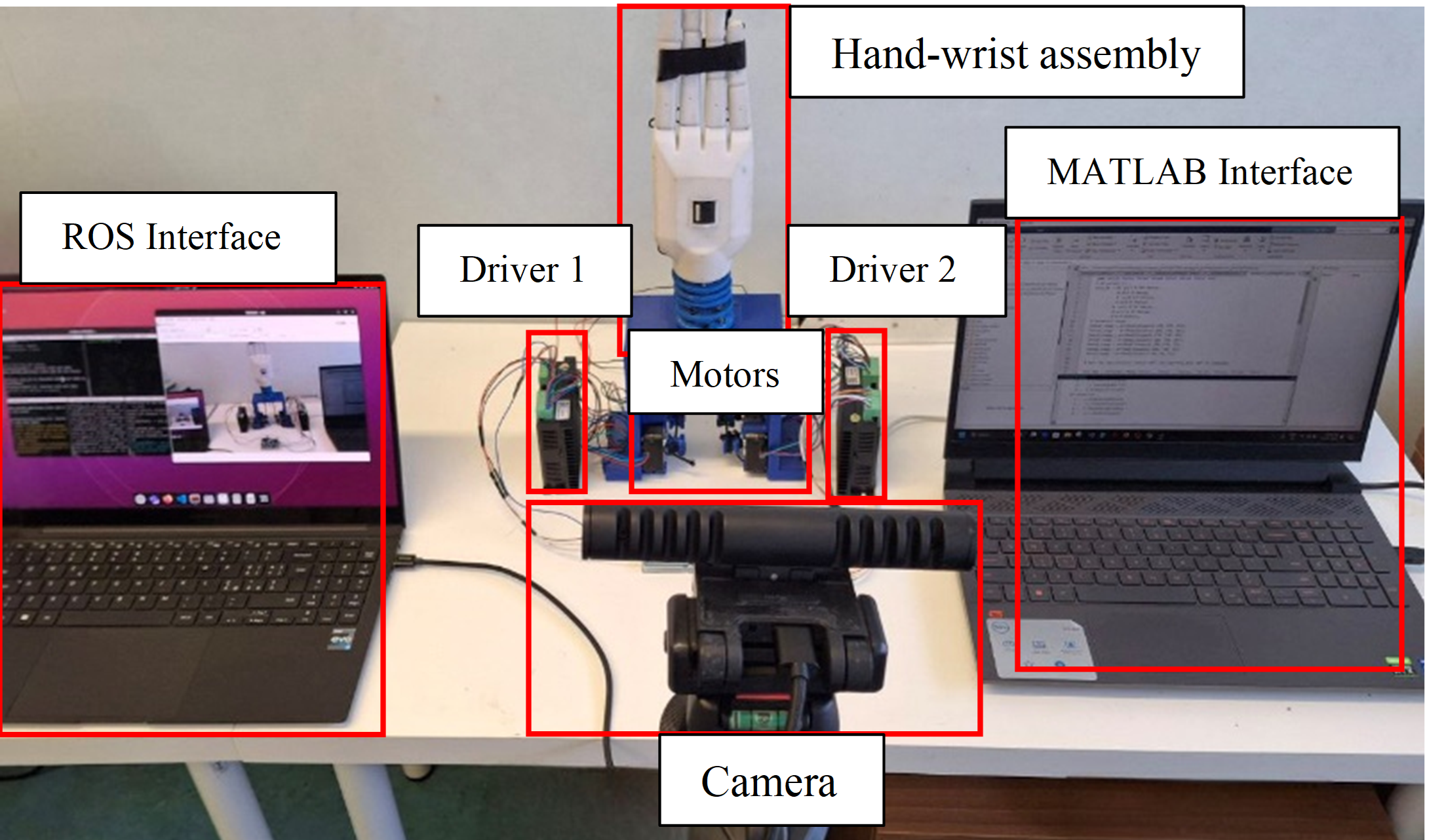}
    \caption{Experimentation set up}
   \label{exp_setup}
\end{figure}

\begin{figure}[hbt!]
    \centering
\includegraphics[width=0.5\textwidth, height = 4 in]{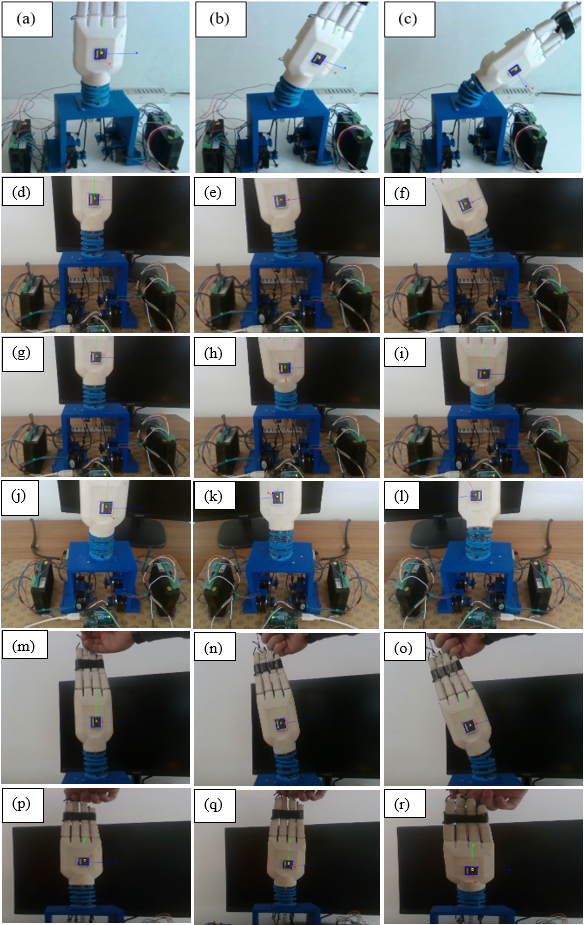} 
    \caption{Motion of the wrist section along with hand (a)-(c)Ulnar (d)-(f)Radial (g)-(i)Extension (j)-(l)Flexion (m)-(o)External disturbance during radial deviation (p)-(r)External disturbance during extension   }
   \label{exp_motions}
\end{figure}

\begin{figure}[hbt!]
    \centering
\includegraphics[width=0.65\textwidth,height =3.8 in]{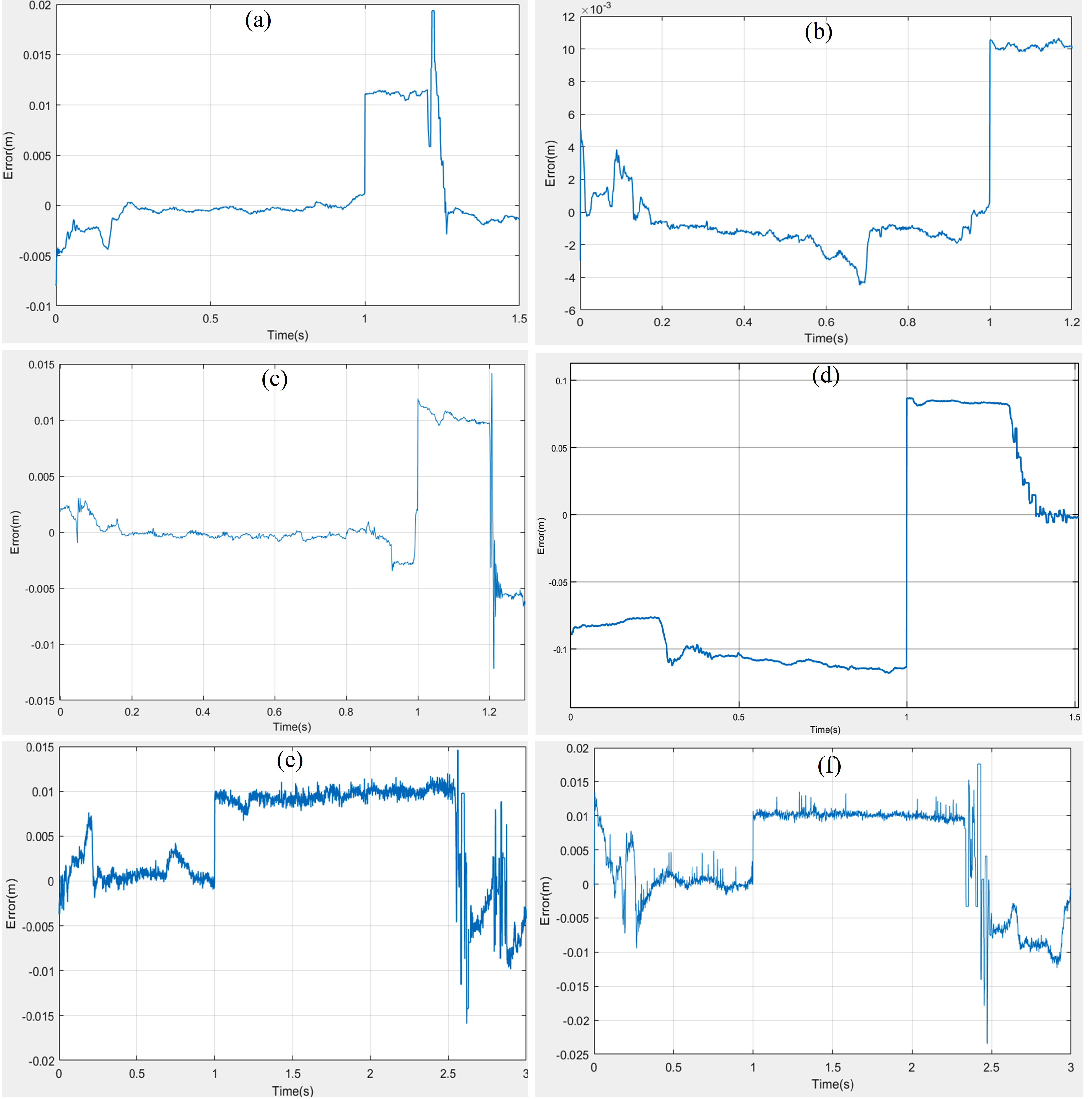}
    \caption{Error during experimentation (a)Ulnar (b)Radial (c)Extension (d)Flexion (e)External disturbance during radial deviation (f)External disturbance during extension }
   \label{exp_errors}
\end{figure}

The experimental setup for the constructed model of the wrist and hand, along with its electronic components, is illustrated in fig. \ref{exp_setup}. An ArUco marker affixed to the hand facilitated the tracking of its poses throughout the experimentation process. The setup included four stepper motors, two motor drivers, a 3D depth camera, and an Arduino controller for real-time operation. Additionally, ROS and MATLAB software were employed for tracking the ArUco poses and executing the control scheme, respectively. Four peripheral tendons were employed to facilitate rotational movements in four distinct directions. Tendons 1 and 2 were activated to achieve radial deviation of the wrist, while tendons 4 and 5 were responsible for movements in the ulnar direction. Additionally, tendons 1 and 4 were utilized to manage extension motions, whereas flexion was controlled by tendons 2 and 5. The lowest disc (disc 1) was affixed to a stable platform, and the highest disc (disc 5) was attached to the hand. Motion of hand in all  directions are shown in fig. \ref{exp_motions} (a) - (l). The errors in motions of the above mentioned motions are shown in fig. \ref{exp_errors} (a) - (d). In the course of the experimentation, the average RMSE values of deflection, settling time, and steady-state error of all directions were measured as $0.05~ m$, $1.35~s$, and $0.031~m$ respectively. The results were considerably higher than those noted in the simulation study. However, the settling time was obtained below 1.5 s demonstrating its computational efficiency during real-time applications. A key reason for the elevated error margin observed in the experimental phase was the reduced stiffness of the springs. In order to validate the adaptiveness of the controller during an application of external unknown forces, we gave an external force using a human hand while the prosthetic hand was moving in radial deviation and flexion directions as shown in fig. \ref{exp_motions} (m) - (r). The errors in motions during application of external disturbances are shown in fig. \ref{exp_errors} (e) - (f). Average RMSE value of the motions during the application of disturbances was obtained as $0.061~ m$. The ability of the controller to function as an adaptive controller has been demonstrated through the achievement of a significantly reduced RMSE value.

\section{Conclusion}
A neuro-controller controller was implemented using a mathematical model determined based on the Timoshenko beam theory during the motion of the wrist section carrying the prosthetic hand. The implementation of the neuro-controller approach enabled the system to sustain the desired motion trajectories despite fluctuations in the robot's physical characteristics and environmental factors. The use of NN enhanced controller performance by decreasing computational time. Furthermore, adaptive control strategy facilitated the faster response. Simulation studies revealed that proposed controller achieved a lower settling time in comparison to other controllers. Additionally, the RMSE and steady-state errors associated with neuro-controller were found to be in the tolerance range. However, during experiments, RMSE values were higher than those observed in simulations, attributed to variations in spring stiffness. Future works will concentrate on redesigning the wrist to enhance structural robustness, while controller strategies will be refined by incorporating real-time sensor feedback to improve motion accuracy.

\section*{Acknowledgement}
We acknowledge funding from Next Generation EU in the context of the National Recovery and Resilience Plan, Investment PE8—Project Age-It: “Ageing Well in an Ageing Society”. This resource was financed by the Next Generation EU (DM 1557 11.10.2022). The views and opinions expressed are only those of the authors and do not necessarily reflect those of the European Union or the European Commission. Neither the European Union nor the European Commission can
be held responsible for them.

\end{document}